\documentclass[10pt,article]{IEEEtran}
\usepackage{import}
\usepackage{amsmath,graphicx}

\usepackage{times}
\usepackage[dvipsnames]{xcolor}

\usepackage{amsmath}

\usepackage{tabularx,booktabs}
\usepackage{arydshln}
\usepackage{tabu}
\usepackage{adjustbox}
\usepackage{enumerate,multicol}
\usepackage{multirow}
\usepackage{multicol}
\usepackage{enumitem}
\usepackage{amsmath,array,graphicx}
\usepackage{gensymb}
\usepackage{float}
\usepackage{caption}
\usepackage{subcaption}
\usepackage{balance}
\usepackage{xspace}
\usepackage{amssymb}
\usepackage{pifont}
\usepackage{booktabs}
\usepackage{subcaption}
\usepackage[pagebackref]{hyperref}

\hypersetup{
    colorlinks,
    linkcolor={red!50!black},
    citecolor={blue!50!black},
    urlcolor={black}
}

\newcommand{\xmark}{\ding{55}}
\newcommand{\ie}{\textit{i.e.}, }

\newcommand{\etal}{\emph{et al.\xspace}}
\usepackage{times}

\usepackage{soul}
\usepackage{url}

\usepackage[utf8]{inputenc}
\usepackage{amsmath}
\usepackage{booktabs}

\usepackage{pgfplots}
\usepackage{pgfplotstable}
\usetikzlibrary{pgfplots.groupplots}
\usepgfplotslibrary{colorbrewer}
\pgfplotsset{/pgfplots/error bars/error bar style={black,thick}}

\pgfplotsset{compat=1.11,
        /pgfplots/ybar legend/.style={
        /pgfplots/legend image code/.code={%
        \draw[##1,/tikz/.cd,bar width=3pt,yshift=-0.2em,bar shift=0pt]
                plot coordinates {(0cm,0.8em)};},
},}

\pgfplotsset{compat=1.7}
\usepackage{ctable}
\definecolor{RYB2}{RGB}{245,245,245}
\definecolor{RYB1}{RGB}{218,232,252}
\definecolor{RYB4}{RGB}{108,142,191}
\newenvironment{customlegend}[1][]{%
    \begingroup
    \csname pgfplots@init@cleared@structures\endcsname
    \pgfplotsset{#1}%
}{%
    \csname pgfplots@createlegend\endcsname
    \endgroup
}%

\def\addlegendimage{\csname pgfplots@addlegendimage\endcsname}
\usepackage{cite}
\usepackage{amsmath,amssymb,amsfonts}
\usepackage{algorithmic}
\usepackage{graphicx}
\usepackage{textcomp}
\def\BibTeX{{\rm B\kern-.05em{\sc i\kern-.025em b}\kern-.08em
    T\kern-.1667em\lower.7ex\hbox{E}\kern-.125emX}}
\begin{document}
\title{DeepPyram: Enabling Pyramid View and Deformable Pyramid Reception for Semantic Segmentation in Cataract Surgery Videos}
\author{Negin Ghamsarian, Mario Taschwer, and klaus Sch\"offmann, \textit{IEEE Member}
\thanks{``This work was funded by the FWF Austrian Science Fund under grant P 31486-N31.'' }
\thanks{Negin Ghamsarian, Mario Taschwer, and Klaus Sch\"offmann are with 
the Information Technology Department, Klagenfurt University, Klagenfurt, Austria (e-mail: \{negin, mt, ks\}@itec.aau.at).}}

\maketitle

\begin{abstract}
Semantic segmentation in cataract surgery has a wide range of applications contributing to surgical outcome enhancement and clinical risk reduction. However, the varying issues in segmenting the different relevant instances make the designation of a unique network quite challenging. This paper proposes a semantic segmentation network termed as DeepPyram that can achieve superior performance in segmenting relevant objects in cataract surgery videos with varying issues. This superiority mainly originates from three modules: (i) Pyramid View Fusion, which provides a varying-angle global view of the surrounding region centering at each pixel position in the input convolutional feature map; (ii) Deformable Pyramid Reception, which enables a wide deformable receptive field that can adapt to geometric transformations in the object of interest; and (iii) Pyramid Loss that adaptively supervises multi-scale semantic feature maps. These modules can effectively boost semantic segmentation performance, especially in the case of transparency, deformability, scalability, and blunt edges in objects.
The proposed approach is evaluated using four datasets of cataract surgery for objects with different contextual features and compared with thirteen state-of-the-art segmentation networks. The experimental results confirm that DeepPyram outperforms the rival approaches without imposing additional trainable parameters. Our comprehensive ablation study further proves the effectiveness of the proposed modules.
\end{abstract}


\section{Introduction}
\label{sec:introduction}
\begin{figure}[!tb]
    \centering
    \includegraphics[width=1\columnwidth]{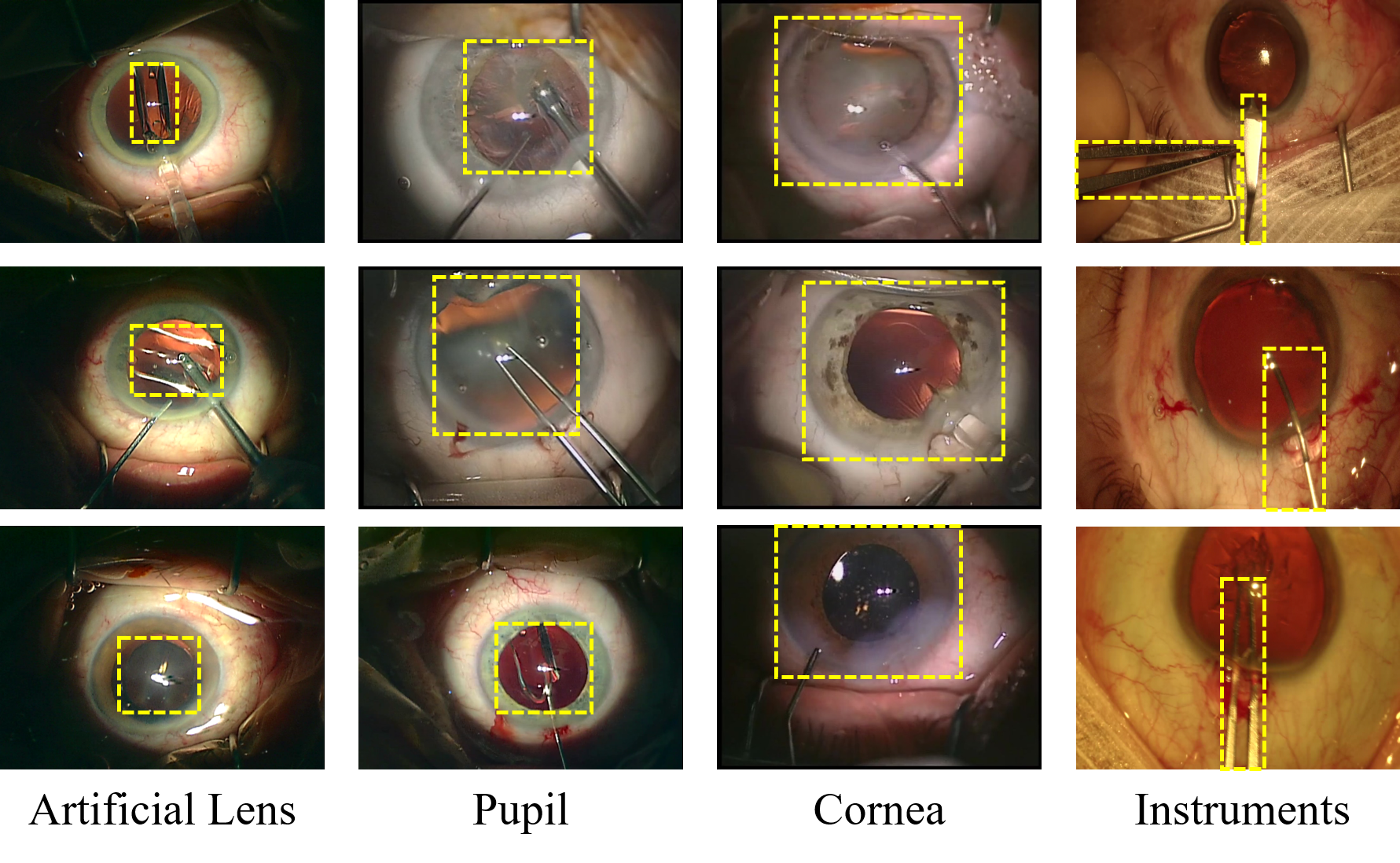}
    \caption{Semantic Segmentation difficulties for different relevant objects in cataract surgery videos.}
    \label{fig: Problems}
\end{figure}

\IEEEPARstart{S}{emantic} segmentation plays a prominent role in computerized surgical workflow analysis. Especially in cataract surgery, where workflow analysis can highly contribute to the reduction of intra-operative and post-operative complications\cite{RBE}, semantic segmentation is of great importance. Cataract refers to the eye's natural lens having become cloudy and causing vision deterioration. Cataract surgery is the procedure of restoring clear eye vision via cataract removal followed by artificial lens implantation. This surgery is the most common ophthalmic surgery and one of the most frequent surgical procedures worldwide~\cite{JFCS}. Semantic segmentation in cataract surgery videos has several applications ranging from phase and action recognition~\cite{RDC, DeepPhase}, irregularity detection (pupillary reaction, lens rotation, lens instability, and lens unfolding delay detection), objective skill assessment, relevance-based compression\cite{RelComp}, and so forth~\cite{IoLP, RTIT, RTSB, NoR, MTU}. Accordingly, there exist four different relevant objects in videos from cataract surgery, namely Intraocular Lens, Pupil, Cornea, and Instruments.  The diversity of features of different relevant objects in cataract surgery imposes a challenge on optimal neural network architecture designation. More concretely, a semantic segmentation network is required that can simultaneously deal with (I) deformability and transparency in case of the artificial lens, (II) color and texture variation in case of the pupil, (III) blunt edges in case of the cornea, and (IV) harsh motion blur degradation, reflection distortion, and scale variation in case of the instruments (Fig.~\ref{fig: Problems}).

\begin{figure}[!tb]
    \centering
    \includegraphics[width=0.75\columnwidth]{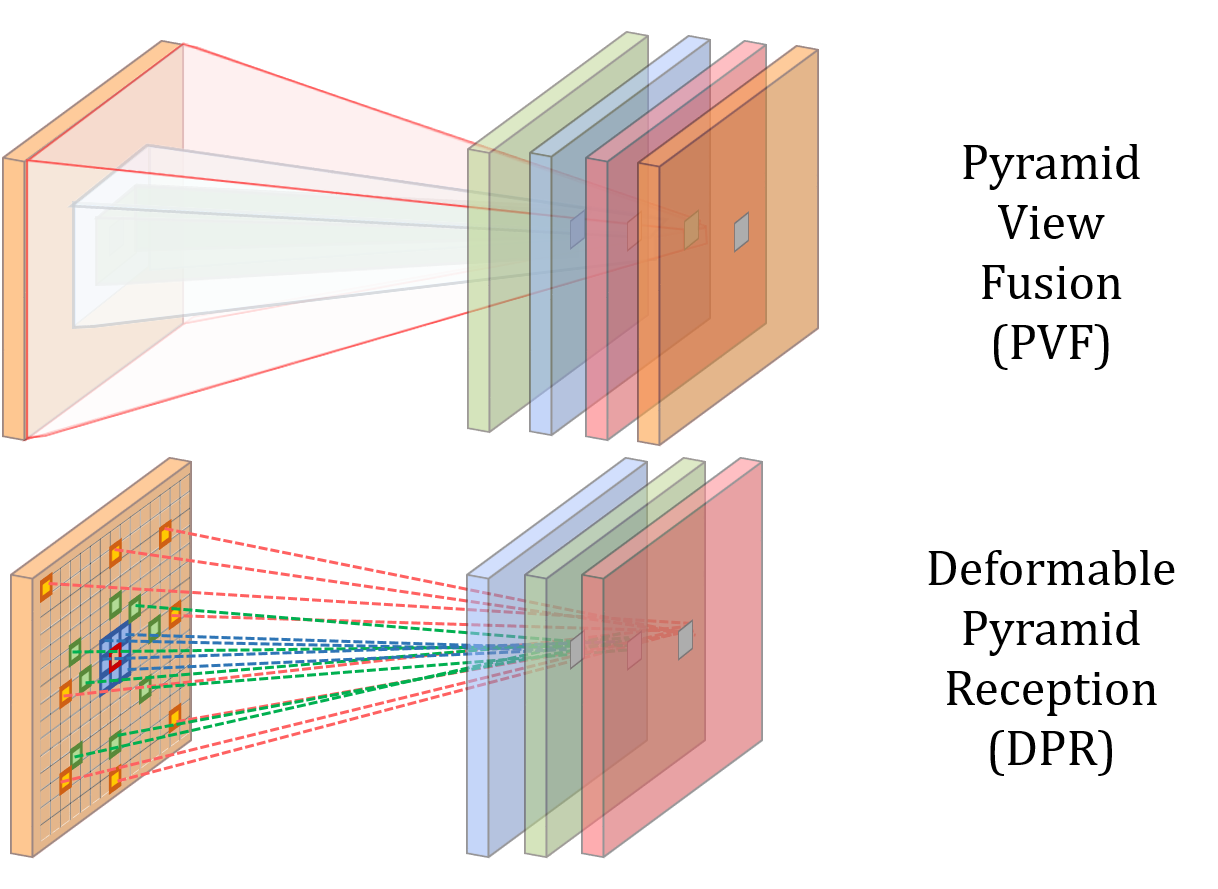}
    \caption{Two major operations in DeepPyram.}
    \label{fig: DeepPyram}
\end{figure}

This paper presents a U-Net-based CNN for semantic segmentation that can adaptively capture the semantic information in cataract surgery videos.
The proposed network, termed as DeepPyram, mainly consists of three modules: (i) Pyramid View Fusion (PVF) module enabling a varying-angle surrounding view of feature map for each pixel position, (ii) Deformable Pyramid Reception (DPR) module being responsible for performing shape-wise feature extraction on the input convolutional feature map  (Fig.~\ref{fig: DeepPyram}), and (iii) Pyramid Loss ($P\mathcal{L}$) module that directly supervises the multi-scale semantic feature maps. We have provided a comprehensive study to compare the performance of DeepPyram with thirteen rival state-of-the-art approaches for relevant-instance segmentation in cataract surgery. The experimental results affirm the superiority of DeepPyram, especially in the case of scalable and deformable transparent objects. To support reproducibility and further comparisons, we will release the PyTorch implementation of DeepPyram and all rival approaches and the customized annotations with the acceptance of this paper.

The rest of the paper is organized as follows. In Section~\ref{sec: relatedwork}, we position our approach in the literature by reviewing the state-of-the-art semantic segmentation approaches. We then delineate the proposed network (DeepPyram) in Section~\ref{sec: Methodology}. We describe the experimental settings in Section~\ref{sec: Experimental Settings} and analyze the experimental results in Section~\ref{sec: Experimental Results}. We also provide an ablation study on DeepPyram in Section~\ref{sec: Experimental Results} and conclude the paper in Section~\ref{sec: Conclusion}.

\begin{figure*}[tb!]
    \centering
    \includegraphics[width=0.85\textwidth]{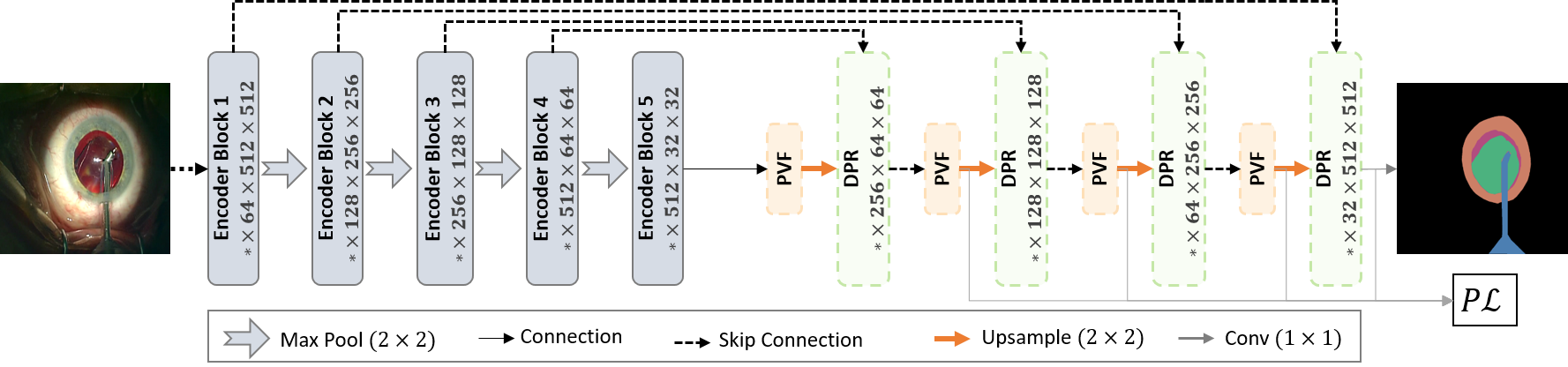}
    \caption{The overall architecture of the proposed DeepPyram network. It contains Pyramid View Fusion (PVF), Deformable Pyramid Reception (DPR), and Pyramid Loss ($P\mathcal{L}$) modules.}
    \label{fig: Block_diagram}
\end{figure*}
\section{Related Work}
\label{sec: relatedwork}

This section briefly reviews U-Net-based approaches, state-of-the-art semantic segmentation approaches related to attention and fusion modules, and multi-branch supervision. 

\vspace{0.5\baselineskip}
\noindent{\textit{\textbf{U-Net-based Networks. }}}
U-Net~\cite{U-Net} was initially proposed for medical image segmentation and achieved succeeding performance being attributed to its skip connections. In the encoder side of this encoder-decoder network, low-level features are combined and transformed into semantic information with low resolution. The decoder network improves these low-resolution semantic features and converts them to semantic segmentation results with the same resolution as the input image. The role of skip connections is to transmit the fine-grained low-level feature maps from the encoder to the decoder layers. The coarse-grained semantic feature maps from the decoder and fine-grained low-level feature maps from the encoder are accumulated and undergo convolutional operations. This accumulation technique helps the decoder retrieve the object of interest's high-resolution features to provide delineated semantic segmentation results. Many U-Net-based architectures have been proposed over the past five years to improve the segmentation accuracy and address different flaws and restrictions in the previous architectures~\cite{SegNet, FED-Net, RAUNet, CE-Net, MultiResUNet, dU-Net, PAANet, BARNet, CPFNet, UNet++}. 
In SegNet~\cite{SegNet}, non-linear upsampling in a decoder's layer is performed using the stored max-pooling indices in its symmetrical encoder's layer. The sparse upsampled feature maps are then converted to dense feature maps using convolutional layers. 
In another work, dU-Net~\cite{dU-Net} replaces the classic convolutional operations in U-Net with deformable convolutions~\cite{DeformConv} to deal with shape variations.
Full-Resolution Residual Networks (FRRN) adopt a two-stream architecture, namely the residual and the pooling stream~\cite{FRRN}. The residual stream preserves the full resolution of the input image. The pooling stream employs consecutive full-resolution residual units (FRRU) to add semantic information to the residual stream gradually.  
Inspired by FRNN, Jue~\etal~\cite{MRRC} developed two types of multiple resolution residually connected networks (MRRN) for lung tumor segmentation. The two versions of MRNN, namely incremental-MRNN and dense-MRNN, aim to fuse varying-level semantic feature maps with different resolutions to deal with size variance in tumors.
It is argued that the optimal depth of a U-Net architecture for different datasets is different. UNet++~\cite{UNet++} as an ensemble of varying-depth UNets is proposed to address this depth optimization problem.
Hejie~\etal~\cite{Fused-UNet++} employe UNet++ as their baseline and add dense connections between the blocks with matching-resolution feature maps for pulmonary vessel segmentation.

\vspace{0.5\baselineskip}
\noindent{\textit{\textbf{Attention Modules. }}}
Attention mechanisms can be broadly described as the techniques to guide the network's computational resources (\ie the convolutional operations) towards the most determinative features in the input feature map. Such mechanisms have been especially proven to be gainful in the case of semantic segmentation.  
Inspired by Squeeze-and-Excitation block~\cite{SAE}, the SegSE block~\cite{AFRR} and scSE block~\cite{SCSE} aim to utilize inter-channel dependencies and recalibrate the channels spatially by applying fully connected operations on the globally pooled feature maps. 
RAUNet~\cite{RAUNet} includes an attention module to merge the multi-level feature maps from the encoder and decoder using global average pooling. BARNet~\cite{BARNet} adopts a bilinear-attention module to extract the cross semantic dependencies between the different channels of a convolutional feature map. This module is specially designed to enhance the segmentation accuracy in the case of illumination and scale variation for surgical instruments. 
PAANET~\cite{PAANet} uses a double attention module (DAM) to model semantic dependencies between channels and spatial positions in the convolutional feature map.

\vspace{0.5\baselineskip}
\noindent{\textit{\textbf{Fusion Modules. }}}
Fusion modules can be characterized as modules designed to improve semantic representation via combining several feature maps. The input feature maps could range from varying-level semantic features to the features coming from parallel operations.
PSPNet~\cite{PSPNet} adopts a pyramid pooling module (PPM) containing parallel sub-region average pooling layers followed by upsampling to fuse the multi-scale sub-region representations. This module has shown significant improvement and is frequently used in semantic segmentation architectures~\cite{3D-PPN}. 
Atrous spatial pyramid pooling (ASPP)~\cite{DeepLab, DeepLabv3+} was proposed to deal with objects' scale variance as an efficient alternative to Share-net~\cite{AtS}. Indeed, the ASPP module aggregates multi-scale features extracted using parallel varying-rate dilated convolutions and obviates the need to propagate and aggregate the features of multi-scale inputs. This module is employed in many segmentation approaches due to its effectiveness in capturing multi-scale contextual features~\cite{MSCA, AFPNet, 3D-ASPP}. 

Autofocus Convolutional Layer~\cite{Autofocus} uses a novel approach to fuse resulting feature maps of dilated convolutions adaptively. CPFNet~\cite{CPFNet} uses another fusion approach for scale-aware feature extraction. MultiResUNet~\cite{MultiResUNet} and Dilated MultiResUNet~\cite{DilatedMultiResUNet} factorize the large and computationally expensive receptive fields into a fusion of successive small receptive fields.
\begin{figure*}[!tb]
    \centering
    \includegraphics[width=1\textwidth]{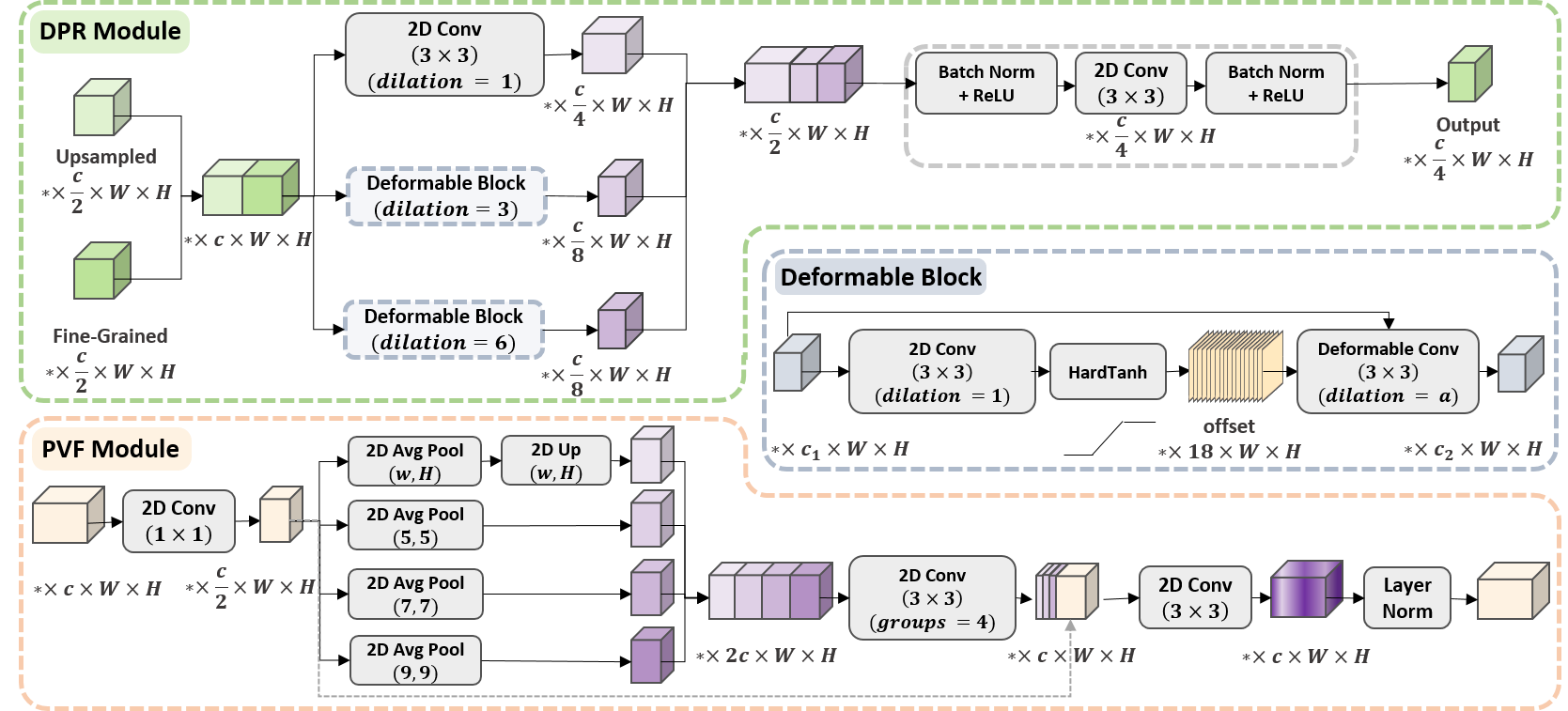}
    \caption{The detailed architecture of the Deformable Pyramid Reception (DPR) and Pyramid View Fusion (PVF) modules.}
    \label{fig: DPR}
\end{figure*}
\vspace{0.5\baselineskip}
\noindent{\textit{\textbf{Multi-Branch Supervision. }}}
The idea of deep supervision was initially proposed by Chen-Yu~\etal~\cite{DSN}. The authors proved that introducing a classifier (SVM or Softmax) on top of hidden layers can improve the learning process and minimize classification failure. 
This idea is simultaneously adopted by GoogleNet~\cite{GoogleNet} to facilitate gradient flow in deep neural network architectures.
The auxiliary loss in PSPNet~\cite{PSPNet} follows the same strategy and guides one feature map in the encoder network to reinforce learning discriminative features in shallower layers.
The architecture of DensNets~\cite{DCCN} implicitly enables such deep supervision. Multi-branch supervision approaches are frequently used for improving semantic segmentation. Qi~\etal~\cite{3DDSN} suggest directly supervising multi-resolution feature maps of the encoder network by adding deconvolutional layers followed by a Softmax activation layer on top of them. Zhu~\etal~\cite{DSPS} adopt five deep supervision modules for the multi-scale feature maps in the encoder network and three deep supervision modules in the decoder network. In each module, the input feature map is upsampled to its original version and undergoes a deconvolutional layer which extracts semantic segmentation results.
The nested architecture of UNet++~\cite{UNet++} inherently provides such multi-depth semantic feature maps with original resolution. 


\section{Methodology}
\label{sec: Methodology}

\subsection{Overview}
Fig.~\ref{fig: Block_diagram} depicts the architecture of the proposed network. Overall, the network consists of a contracting path and an expanding path. The contracting path is responsible for converting low-level to semantic features. The expanding path accounts for performing super-resolution on the coarse-grained semantic feature maps and improving the segmentation accuracy by taking advantage of the symmetric\footnote{Symmetric feature maps are the feature maps with the same spatial resolution.} fine-grained feature maps. The encoder network in the baseline approach is VGG16, pretrained on ImageNet. The decoder network consists of three modules: (i) Pyramid View Fusion (PVF), which induces a large-scale view with progressive angles, (ii) Deformable Pyramid Reception (DPR), which enables a large, sparse, and learnable receptive field, which can sample from up to seven pixels far from each pixel position in the input convolutional feature map, and (iii) Pyramid Loss ($P\mathcal{L}$), which is responsible for the direct supervision of multi-scale semantic feature maps. In the following subsections, we detail each of the proposed modules.

\subsection{Pyramid View Fusion (PVF)} By this module, we aim to stimulate a neural network deduction process analogous to the human visual system. Due to non-uniformly distributed photoreceptors on the retina, the perceived region with high resolution by the human visual system is up to $2-5\degree$ of visual angle centering around the gaze~\cite{Af}. Correspondingly, we infer that the human eye recognizes the semantic information considering not only the internal object's content but also the relative information between the object and the surrounding area. The pyramid view fusion (PVF) module's role is to reinforce the feeling of such relative information at every distinct pixel position. One way to exploit such relative features is to apply convolutional operations with large receptive fields. However, increasing the receptive field's size is not recommended due to imposing huge additional trainable parameters, and consequently (i) escalating the risk of overfitting and (ii) increasing the requirement to more annotations. Alternatively, we use average pooling for fusing the multi-angle local information in our novel attention mechanism. At first, as shown at the bottom of Fig.~\ref{fig: DPR}, a bottleneck is formed by employing a convolutional layer with a kernel size of one to curb the computational complexity. After this dimensionality reduction stage, the convolutional feature map is fed into four parallel branches. The first branch is a global average pooling layer followed by upsampling. The other three branches include average pooling layers with progressive filter sizes and the stride of one. Using a one-pixel stride is specifically essential for obtaining pixel-wise centralized pyramid view in contrast with region-wise pyramid attention in PSPNet~\cite{PSPNet}. The output feature maps are then concatenated and fed into a convolutional layer with four groups. This layer is responsible for extracting inter-channel dependencies during dimensionality reduction. A regular convolutional layer is then applied to extract joint intra-channel and inter-channel dependencies before being fed into a layer-normalization function.

\vspace{0.5\baselineskip}
\noindent{\textit{\textbf{Discussion. }}}There are three significant differences between the \textbf{PVF module} in DeepPyram and pyramid pooling module in PSPNet~\cite{PSPNet}: (i) All average pooling layers in the PVF module use a stride of one pixel, whereas the average pooling layers in PSPNet adopt different strides of 3, 4, and 6 pixels. The pyramid pooling module separates the input feature map into varying-size sub-regions and plays the role of object detection as region proposal networks (RPNs~\cite{RPN}). However, the stride of one pixel in the PVF module is used to capture the subtle changes in pyramid information that is especially important for segmenting the narrow regions of objects such as instruments and the artificial lens hooks\footnote{Precise segmentation of the artificial lens hooks is crucial for lens rotation and irregularity detection.}. These subtle differences can be diluted after fusing the pyramid information in the pyramid pooling module in PSPNet. (ii) In contrast with PSPNet, the PVF module applies three functions to the concatenated features to capture high-level contextual features: (1) a group convolution function to fuse the features obtained from the average pooling filters independently (as in PSPNet), (2) a convolutional operation being responsible for combining the multi-angle local features to reinforce the semantic features corresponding to each target label, and (3) a layer normalization function that accelerates training and avoids overfitting to the color and contrast information in case of few annotations.  

\subsection{Deformable Pyramid Reception (DPR)}
Fig.~\ref{fig: DPR} (top) demonstrates the architecture of the deformable pyramid reception (DPR) module. At first, the fine-grained feature map from the encoder and coarse-grained semantic feature map from the previous layer are concatenated. Afterward, these features are fed into three parallel branches: a regular $3\times 3$ convolution and two deformable blocks with different dilation rates. These layers together cover a learnable sparse receptive field of size $15\times 15$\footnote{The structured $3\times 3$ filter covers up to one pixel far from the central pixel). The deformable filter with $dilation=3$ covers an area from two to four pixels far away from each central pixel. The second deformable convolution with $dilation=6$ covers an area from five to seven pixels far away from each central pixel. Therefore, these layers together form a sparse filter of size $15\times 15$ pixels. This sparse kernel can be better seen in Fig.~\ref{fig: DeepPyram}.}. The output feature maps are then concatenated before undergoing a sequence of regular layers for higher-order feature extraction and dimensionality reduction.

\vspace{0.5\baselineskip}
\noindent{\textit{\textbf{Deformable Block. }}}
Dilated convolutions can implicitly enlarge the receptive field without imposing additional trainable parameters. Dilated convolutional layers can recognize each pixel position's semantic label based on its cross-dependencies with varying-distance surrounding pixels. These layers are exploited in many architectures for semantic segmentation~\cite{MSCA, DeepLab, AFPNet}. Due to the inflexible rectangle shape of the receptive field in regular convolutional layers, however,  the feature extraction procedure cannot be adapted to the target semantic label's shape. Dilated deformable convolutional layers can effectively support the object's geometric transformations in terms of scale and shape. As shown in the \textit{Deformable Block} in Fig.~\ref{fig: DPR}, a regular convolutional layer is applied to the input feature map to compute the offset field for deformable convolution. The offset field provides two values per element in the convolutional filter (horizontal and vertical offsets). Accordingly, the number of offset field's output channels for a kernel of size $3\times 3$ is equal to 18. Inspired by dU-Net~\cite{dU-Net}, the convolutional layer for the offset field is followed by an activation function. We use the hard tangent hyperbolic function (HardTanh), which is computationally cheap, to clip the offset values in the range of $[-1,1]$. The deformable convolutional layer uses the learned offset values along with the convolutional feature map with a predetermined dilation rate to extract object-adaptive features.

The output feature map ($y$) for each pixel position ($p_0$) and the receptive field ($\mathcal{RF}$) for a regular 2D convolution with a $3\times 3$ filter and dilation rate of one can be defined as:

\begin{equation}
    y(p_o)=\sum_{p_i\in \mathcal{RF}_1}{x(p_0+p_i).w(p_i)}
\end{equation}
\begin{equation}
    \mathcal{RF}_1=\{(-1,-1),(-1,0), ..., (1,0),(1,1)\}
\end{equation}

Where $x$ denotes the input convolutional feature map and $w$ refers to the weights of the convolutional kernel. In a dilated 2D convolution with a dilation rate of $\alpha$, the receptive field can be defined as $\mathcal{RF}_{\alpha}=\alpha \times \mathcal{RF}_1$. Although the sampling locations in a dilated receptive field have a greater distance with the central pixel, they follow a firm structure. In a deformable dilated convolution with a dilation rate of $\alpha$, the sampling locations of the receptive field are dependent to the local contextual features. In the proposed deformable block, the sampling location for the $i$th element of the receptive field and the input pixel $p_0$ can be formulated as:

\begin{equation}
\begin{aligned}
   \mathcal{RF}_{def,\alpha}[i,p_{0}]=& \mathcal{RF}_\alpha[i]\\
   +&f(\sum_{p_j\in \mathcal{RF}_1}{x(p_0+p_j).\hat{w}(p_j)})
\end{aligned}   
\label{eq: RF3}
\end{equation}

In~\eqref{eq: RF3}, $f$ denotes the activation function, which is the tangent hyperbolic function in our case, and $\hat{w}$ refers to the weights of the offset filter. This learnable receptive field can be adapted to every distinct pixel in the convolutional feature map and allows the convolutional layer to extract more informative semantic features compared to the regular convolution.
\begin{figure}[!tb]
    \centering
    \includegraphics[width=1\columnwidth]{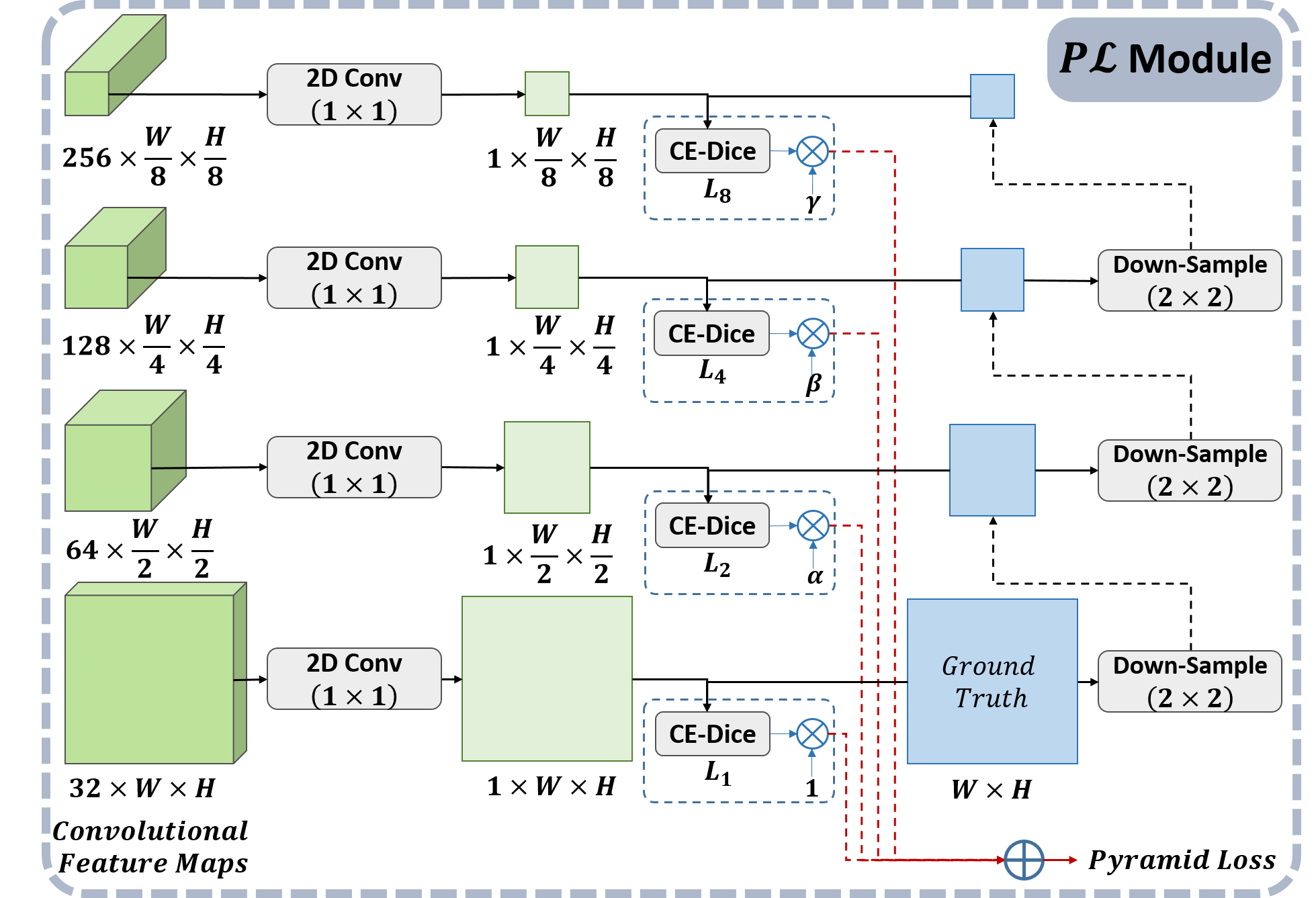}
    \caption{Demonstration of the \textit{Pyramid Loss} module.}
    \label{fig:PL}
\end{figure}

\vspace{0.5\baselineskip}
\noindent{\textit{\textbf{Discussion. }}}deformable convolutions are usually used for video object detection, tracking, and segmentation since a deformable layer needs offsets. These offsets are usually provided by optical flow computation, subtracting consecutive frames, or substracting the corresponding feature maps of consecutive frames (as in MaskProp~\cite{MaskProp} and MF-TAPNet~\cite{ITP}). Since we use videos recorded in usual and non-laboratory conditions, we have many problems in the video dataset such as defocus blur and harsh motion blur (due to fast movements of eye and motions of the instruments~\cite{DCS}). Using temporal information for semantic segmentation (as in MaskProp~\cite{MaskProp}) may lead to error accumulation and less precise results. Moreover, video object segmentation requires much more annotations which is an additional burden on the expert surgeons' time. For the DPR module, we propose (1) a deformable block to apply deformable convolutions based on learned offsets from static information, (2) a combination of deformable and static filters which can not only capture the edge-sensitive information in the case of sharp edges (as in pupil and lens), but also can cover a large area (up to a size of $15\times 15$ pixels) to deal with color and scale variations, blunt edges in case of cornea, and reflections in case of instruments.

\subsection{Pyramid Loss (P$\mathcal{L}$)}
The role of this module is to directly supervise the multi-scale semantic feature maps on the decoder's side. As shown in Fig~\ref{fig:PL}, in order to enable direct supervision, a fully connected layer is formed using a pixel-wise convolution operation. The output feature map presents the semantic segmentation results with the same resolution as the input feature map. To compute the loss for varying-scale outputs, we downscale the ground-truth masks using inter-nearest downsampling for multi-class segmentation and max-pooling for binary segmentation. The overall loss is defined as:

\begin{equation}
    P\mathcal{L} = \mathcal{L}_1 + \alpha \mathcal{L}_2 + \beta \mathcal{L}_4 + \gamma \mathcal{L}_8
\end{equation}

Where $\alpha$, $\beta$, and $\gamma$ are predetermined weights in the range of $[0,1]$ (In the experiments, we have set $\alpha = 0.75$, $\beta=0.5$, and $\gamma = 0.25$). Besides, $\mathcal{L}_i$ denotes the loss of output mask segmentation result with the resolution of $(1/i)$ compared to the input resolution.

\vspace{0.5\baselineskip}
\noindent{\textit{\textbf{Discussion. }}}The auxiliary losses for deep semantic segmentation that are proposed to date have two different purposes: (1) Many auxiliary losses aim to guide one or more feature maps in the encoder subnetwork to prevent gradient vanishing due to using a deep architecture (for instance ResNet50) as the backbone. (2) Some auxiliary losses attempt to improve the segmentation accuracy by directly guiding different layers of the encoder or decoder network. In the second case, the loss is computed by performing super-resolution through interpolating the output feature map (to obtain a feature map with the same resolution as the ground truth) and comparing it to the ground truth.  In contrast, the $P\mathcal{L}$ module directly compares each feature map to the downsampled version of the ground truth. This strategy is more time-efficient and computationally less expensive compared to the previously proposed auxiliary losses. Moreover, in contrast with state-of-the-art approaches, each loss branch in the $P\mathcal{L}$ module only consists of $Conv2D + Softmax$ to keep the number of trainable parameters as few as possible.

\begin{table*}[th!]
\renewcommand{\arraystretch}{0.9}
\caption{Specifications of the proposed and rival approaches. In the ``loss" column, ``CE" and ``CE-Dice" stand for \textit{Cross Entropy} and \textit{Cross Entropy Log Dice}. In ``Upsampling" column, ``Trans Conv" stands for \textit{Transposed Convolution}.}

\label{tab:specification}
\centering
\begin{tabularx}{0.90\textwidth}{lccccccc}
\specialrule{.12em}{.05em}{.05em}
Model & Backbone & Params & Loss & Upsampling & Target & Year & Reference\\\specialrule{.12em}{.05em}{.05em}
UNet$++$~&VGG16&24.24 M& CE-Dice & Bilinear & Medical Images & 2020 & ~\cite{UNet++}\\
UNet++\slash DS &VGG16& 24.24 M & CE-Dice & Bilinear & Medical Images & 2020 & ~\cite{UNet++}\\
CPFNet & ResNet34 &34.66 M& CE-Dice & Bilinear & Medical Images & 2020 & \cite{CPFNet}\\
BARNet&ResNet34&24.90 M& CE-Dice & Bilinear & Surgical
Instruments  & 2020 & ~\cite{BARNet}\\
PAANet &ResNet34& 22.43M & CE-Dice & Trans Conv \& Bilinear & Surgical Instruments & 2020 & \cite{PAANet}\\
dU-Net&\xmark &31.98 M& CE-Dice &Trans Conv & Blood Cells & 2020 & ~\cite{dU-Net}\\
MultiResUNet &\xmark& 9.34 M& CE & Trans Conv & Medical Images & 2020 & ~\cite{MultiResUNet}\\
CE-Net &ResNet34&29.9 M&CE&Trans Conv & Medical Images & 2019 & \cite{CE-Net}\\
RAUNet&ResNet34&22.14 M&CE-Dice&Trans Conv& Cataract  Surgical  Instruments & 2019 &~\cite{RAUNet}\\
FED-Net&ResNet50&59.52 M&CE-Dice& Trans Conv \& PixelShuffle & Liver  Lesion & 2019 &~\cite{FED-Net}\\
PSPNet&ResNet50&22.26 M&CE&Bilinear& Scene & 2017 &~\cite{PSPNet}\\
SegNet&VGG16&14.71 M&CE&Max Unpooling& Scene & 2017 &~\cite{SegNet}\\
U-Net&\xmark &17.26 M&CE& Bilinear & Medical Images & 2015 &~\cite{U-Net}\\\cdashline{1-8}[0.8pt/1pt]
DeepPyram&VGG16 &23.62 M&CE-Dice& Bilinear & Cataract Surgery &\multicolumn{2}{c}{Proposed Approach}\\
\specialrule{.12em}{.05em}{0.05em}
\end{tabularx}

\end{table*}
\section{Experimental Setup}
\label{sec: Experimental Settings}

\noindent{\textit{\textbf{Datasets. }}}
We have used four datasets with varying instance features to provide extensive evaluations for the proposed and rival approaches. The two public datasets include the ``Cornea''~\cite{RelComp} and ``Instruments"~\cite{CaDIS} mask annotations. Additionally, we have prepared a customized dataset for ``Intraocular Lens" and ``Pupil"  pixel-wise segmentation\footnote{The dataset will be publicly released with the acceptance of the paper.}. The number of training and testing images for the aforementioned objects are 178:84, 3190:459, 141:48, and 141:48, respectively. All training and testing images are sampled from distinctive videos to meet real-world conditions. Our annotations are performed using the ``supervise.ly" platform\footnote{https://supervise.ly/} based on the guidelines from cataract surgeons.

\vspace{0.5\baselineskip}
\noindent{\textit{\textbf{Rival Approaches. }}}
Table~\ref{tab:specification} details the specifications of the proposed approach and rival approaches employed in our experiments. To provide a fair comparison, we adopt our improved version of PSPNet, featuring a decoder designed similarly to U-Net (with four sequences of double-convolution blocks). Besides, our version of du-Net has the same number of filter-response-maps as for U-Net.

\vspace{0.5\baselineskip}
\noindent{\textit{\textbf{Data Augmentation Methods. }}}
Data augmentation is a vital step during training, which prevents network overfitting and boosts the network performance in the case of unseen data. Accordingly, the training images for all evaluations undergo various augmentation approaches before being fed into the network.  We chose the transformations considering the inherent and statistical features of datasets. For instance, we use motion blur transformation to encourage the network to deal with harsh motion blur regularly occurring in cataract surgery videos~\cite{DCS}.  Table~\ref{tab:aug} details the adopted augmentation pipeline. Each listed method is randomly applied to the input pairs of images and masks with the probability of $0.5$.

\begin{table}[th!]
\renewcommand{\arraystretch}{0.9}
\caption{Augmentation Pipeline.}
\label{tab:aug}
\centering
\begin{tabu}{lcc}
\specialrule{.12em}{.05em}{.05em}
Augmentation Method&Property&Value\\\specialrule{.12em}{.05em}{.05em}
Brightness~\& Contrast&Factor Range&(-0.2,0.2)\\
Shift~\& Scale&Percentage&10\%\\
Rotate&Degree Range&[-10,10]\\
Motion Blur&Kernel-Size Range &(3,7)\\
\specialrule{.12em}{.05em}{0.05em}
\end{tabu}
\end{table}

\vspace{0.5\baselineskip}
\noindent{\textit{\textbf{Neural Network Settings. }}}
As listed in Table~\ref{tab:specification}, U-Net, MultiResUNet, and dU-Net do not adopt a pretrained backbone. For the other approaches, the weights of the backbone are initialized with ImageNet~\cite{ImageNet} training weights.
The input size of all models is set to $3\times 512\times 512$. In the evaluation and testing stages of UNet++/DS and DeepPyram, we disregard the additional output branches (branches related to auxiliary loss functions) and only consider the master output branch.

\vspace{0.5\baselineskip}
\noindent{\textit{\textbf{Training Settings. }}}
Due to the different depth and connections of the proposed and rival approaches, all networks are trained with three different initial learning rates ($lr\in\{0.0005,0.0002,0.001\}$), and the results with the highest IoU for each network are listed. The learning rate is scheduled to decrease every two epochs with the factor of $0.8$. In all different evaluations, the networks are trained end-to-end and for 30 epochs. We use a threshold of $0.1$ for gradient clipping during training\footnote{Gradient clipping is used to clip the error derivatives during back-propagation to prevent gradient explosion.}.

\vspace{0.5\baselineskip}
\noindent{\textit{\textbf{Loss Function. }}}
The \textit{cross entropy log dice} loss, which is used during training, is a weighted sum of binary cross-entropy ($BCE$) and the logarithm of soft Dice coefficient as follows:

\begin{equation}
\begin{aligned}
    \mathcal{L} = &(\lambda)\times BCE(\mathcal{X}_{true}(i,j),\mathcal{X}_{pred}(i,j))\\
    &-(1-\lambda)\times (\log \frac{2\sum \mathcal{X}_{true}\odot \mathcal{X}_{pred}+\sigma}{\sum \mathcal{X}_{true} + \sum \mathcal{X}_{pred}+ \sigma})
\end{aligned}
\label{eq: loss}
\end{equation}

Where $\mathcal{X}_{true}$ denote the ground truth binary mask, and $\mathcal{X}_{pred}$ denote the predicted mask ($0\leq \mathcal{X}_{pred}(i,j) \leq 1$). The parameter $\lambda \in [0,1]$ is set to $0.8$ in our experiments, and $\odot$ refers to Hadamard product (element-wise multiplication). Besides, the parameter $s$ is the Laplacian smoothing factor which is added to (i) prevent division by zero, and (ii) avoid overfitting (in experiments, $\sigma = 1$).

\vspace{0.5\baselineskip}
\noindent{\textit{\textbf{Evaluation Metrics. }}}
The Jaccard metric (Intersection-over-Union -- IoU) and the Dice Coefficient (F1-score) are regarded as the major semantic segmentation indicators. Accordingly, we evaluate the proposed and rival approaches using average IoU and dice. In order to enable a broader analysis of the networks' performance, the standard deviation of IoU, and minimum and maximum of dice coefficient over all of the testing images are additionally compared.

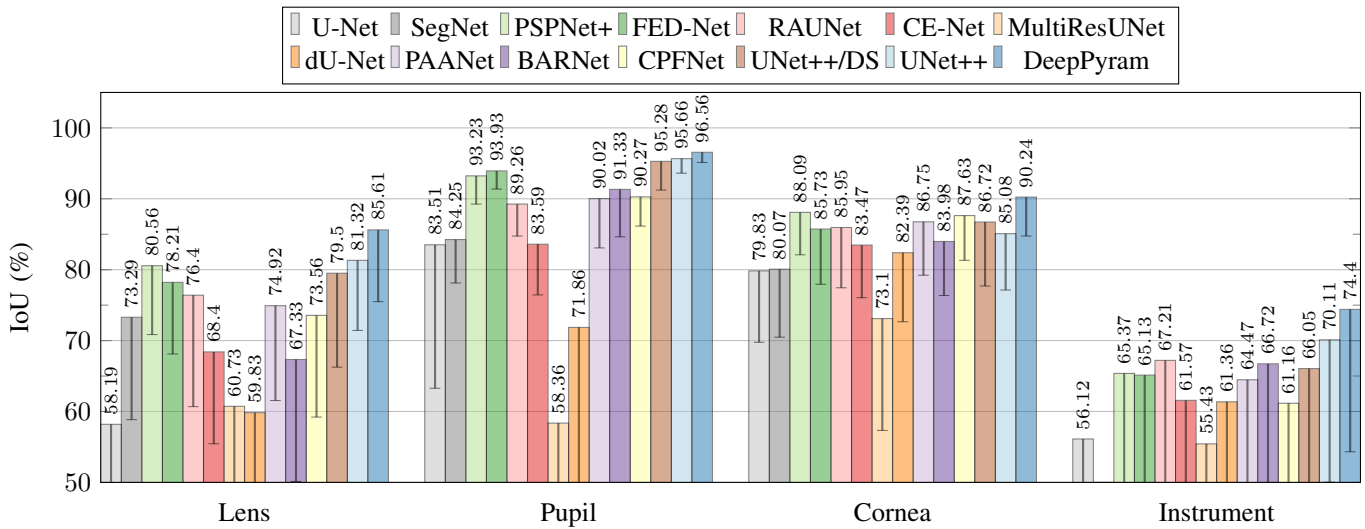
\begin {figure*}
\begin{adjustbox}{width=1\textwidth}

\begin{tikzpicture}
    \begin{axis}[
            ybar=0pt,
            bar width=0.27cm,
            width=\textwidth,
            height=.37\textwidth,
            legend style={at={(0.5,1.22
            )},
                anchor=north,legend columns=7},
            symbolic x coords={Lens,Pupil,Cornea, Instrument},
            xtick=data,
            nodes near coords align={vertical},
            ymin=50,ymax=105,
            ylabel= IoU (\%),
            enlarge x limits={abs=7*\pgfplotbarwidth},
            minor ytick={65,70,75,80,85},
            major x tick style = transparent,
            ymajorgrids = true,
            cycle list/Paired,
            nodes near coords,
            every node near coord/.append style={rotate=90, anchor=west, font=\scriptsize, color = black, opacity=1}
        ]
        \addplot+[style={draw=black,fill=lightgray,opacity=0.5}, 
             error bars/.cd, 
             y dir=minus,y explicit]
             coordinates {
                  (Lens,58.19) +- (0, 30.35)
                  (Pupil,83.51) +- (0.0, 20.24)
                  (Cornea,79.83) +- (0.0, 10.07)
                  (Instrument,56.12) +- (0.0, 27.54)};

        \addplot+[style={draw=black,solid,fill=gray,opacity=0.5}, 
             error bars/.cd, 
             y dir=minus,y explicit]
             coordinates {
                  (Lens,73.29) +- (0.0, 14.44)
                  (Pupil,84.25) +- (0.0, 6.13)
                  (Cornea,80.07) +- (0.0, 9.59)
                  (Instrument,0) +- (0.0, 0)};            
        \addplot+[style={draw=black,solid,fill,opacity=0.5}, 
             error bars/.cd, 
             y dir=minus,y explicit]
             coordinates {
                  (Lens,80.56) +- (0.0, 9.71)
                  (Pupil,93.23) +- (0.0, 3.98)
                  (Cornea,88.09) +- (0.0, 5.99)
                  (Instrument,65.37) +- (0.0, 21.39)};         
                  
        \addplot+[style={draw=black,solid,fill,opacity=0.5}, 
             error bars/.cd, 
             y dir=minus,y explicit]
             coordinates {
                  (Lens,78.21) +- (0.0, 10.10)
                  (Pupil,93.93) +- (0.0, 2.56)
                  (Cornea,85.73) +- (0.0, 7.79)
                  (Instrument,65.13) +- (0.0, 22.66)};         
                  
        \addplot+[style={draw=black,solid,fill,opacity=0.5}, 
             error bars/.cd, 
             y dir=minus,y explicit]
             coordinates {
                  (Lens,76.40) +- (0.0, 15.72)
                  (Pupil,89.26) +- (0.0, 4.52)
                  (Cornea,85.95) +- (0.0, 8.50)
                  (Instrument,67.21) +- (0.0, 20.84)};
                  
        \addplot+[style={draw=black,solid,fill,opacity=0.5}, 
             error bars/.cd, 
             y dir=minus,y explicit]
             coordinates {
                  (Lens,68.40) +- (0.0, 12.95)
                  (Pupil,83.59) +- (0.0, 7.15)
                  (Cornea,83.47) +- (0.0, 7.42)
                  (Instrument,61.57) +- (0.0, 16.77)};

        \addplot+[style={draw=black,solid,fill,opacity=0.5}, 
             error bars/.cd, 
             y dir=minus,y explicit]
             coordinates {
                  (Lens,60.73) +- (0.0, 25.84)
                  (Pupil,58.36) +- (0.0, 33.80)
                  (Cornea,73.10) +- (0.0, 15.77)
                  (Instrument,55.43) +- (0.0, 28.44)};

        \addplot+[style={draw=black,solid,fill,opacity=0.5}, 
             error bars/.cd, 
             y dir=minus,y explicit]
             coordinates {
                  (Lens,59.83) +- (0.0, 29.41)
                  (Pupil,71.86) +- (0.0, 28.88)
                  (Cornea,82.39) +- (0.0, 9.73)
                  (Instrument,61.36) +- (0.0, 27.21)};   
                  
        \addplot+[style={draw=black,solid,fill,opacity=0.5}, 
             error bars/.cd, 
             y dir=minus,y explicit]
             coordinates {
                  (Lens,74.92) +- (0.0, 13.37)
                  (Pupil,90.02) +- (0.0, 6.95)
                  (Cornea,86.75) +- (0.0, 7.53)
                  (Instrument,64.47) +- (0.0, 23.19)};

        \addplot+[style={draw=black,solid,fill,opacity=0.5},  
             error bars/.cd, 
             y dir=minus,y explicit]
             coordinates {
                  (Lens,67.33) +- (0.0, 17.20)
                  (Pupil,91.33) +- (0.0, 6.69)
                  (Cornea,83.98) +- (0.0, 7.62)
                  (Instrument,66.72) +- (0.0, 22.91)};   
                  
        \addplot+[style={draw=black,solid,fill,opacity=0.5}, 
             error bars/.cd, 
             y dir=minus,y explicit]
             coordinates {
                  (Lens,73.56) +- (0.0, 14.35)
                  (Pupil,90.27) +- (0.0, 4.12)
                  (Cornea,87.63) +- (0.0, 6.31)
                  (Instrument,61.16) +- (0.0, 20.12)};          
                  
        \addplot+[style={draw=black,solid,fill,opacity=0.5},  
             error bars/.cd, 
             y dir=minus,y explicit]
             coordinates {
                  (Lens,79.50) +- (0.0, 13.25)
                  (Pupil,95.28) +- (0.0, 4.05)
                  (Cornea,86.72) +- (0.0, 9.02)
                  (Instrument,66.05) +- (0.0, 24.91)};          
                  
        \addplot+[style={draw=black,solid,fill,opacity=0.5},  
             error bars/.cd, 
             y dir=minus,y explicit]
             coordinates {
                  (Lens,81.32) +- (0.0, 9.89)
                  (Pupil,95.66) +- (0.0, 2.05)
                  (Cornea,85.08) +- (0.0,7.95)
                  (Instrument,70.11) +- (0.0, 20.07)};            
                  
        \addplot+[style={draw=black,solid,fill,opacity=0.5}, 
             error bars/.cd, 
             y dir=minus,y explicit]
             coordinates {
                  (Lens,85.61) +- (0.0, 10.12)
                  (Pupil,96.56) +- (0.0, 1.44)
                  (Cornea,90.24) +- (0.0, 5.49)
                  (Instrument,74.40) +- (0.0, 20.08)}; 
                  
        \legend{U-Net, SegNet, PSPNet+, FED-Net, RAUNet, CE-Net, MultiResUNet, dU-Net, PAANet, BARNet, CPFNet, UNet++\slash DS, UNet++, DeepPyram}
    \end{axis}
\end{tikzpicture}

\end{adjustbox}
\caption{Quantitative comparisons among DeepPyram and rival approaches based on average and standard deviation of IoU.}
\label{IoU}
\end{figure*}
\begin{table*}[t!]
\renewcommand{\arraystretch}{0.9}
\caption{Impact of different modules on the segmentation results (IoU\% and Dice\%) of DeepPyram.}
\label{tab:ablation2}
\centering

\begin{tabular}{m{0.5cm} m{0.5cm} m{0.5cm} ccccccccccc}
\specialrule{.12em}{.05em}{.05em}
\multicolumn{3}{c}{Modules}&&\multicolumn{2}{c}{Lens}&\multicolumn{2}{c}{Pupil}&\multicolumn{2}{c}{Cornea}&\multicolumn{2}{c}{Instrument}&\multicolumn{2}{c}{Overall}\\\cmidrule(lr){1-3}\cmidrule(lr){5-6}\cmidrule(lr){7-8}\cmidrule(lr){9-10}\cmidrule(lr){11-12}\cmidrule(lr){13-14}
PVF&DPR&$P\mathcal{L}$&Params&IoU(\%)&Dice(\%)&IoU(\%)&Dice(\%)&IoU(\%)&Dice(\%)&IoU(\%)&Dice(\%)&IoU(\%)&Dice(\%)\\\specialrule{.12em}{.05em}{.05em}
\xmark&\xmark&\xmark&22.55 M&82.98&90.44&95.13&97.48&86.02&92.28&69.82&79.05&83.49&89.81\\
\checkmark&\xmark&\xmark&22.99 M&83.73&90.79&96.04&97.95&88.43&93.77&72.58&81.84&85.19&91.09\\
\xmark&\checkmark&\xmark&23.17 M&81.85&89.58&95.32&97.59&86.43&92.55&71.57&80.60&83.79&90.08\\
\checkmark&\checkmark&\xmark&23.62 M&83.85&90.89&95.70&97.79&89.36&94.29&72.76&82.00&85.42&91.24\\
\checkmark&\checkmark&\checkmark&23.62 M&\textbf{85.84}&\textbf{91.98}&\textbf{96.56}&\textbf{98.24}&\textbf{90.24}&\textbf{94.77}&\textbf{74.40}&\textbf{83.30}&\textbf{86.76}&\textbf{92.07}\\ \specialrule{.12em}{.05em}{.05em}
\end{tabular}

\label{tab:modules}
\end{table*}

\vspace{0.5\baselineskip}
\noindent{\textit{\textbf{Ablation Study Settings. }}}
To evaluate the effectiveness of different modules, we have implemented another segmentation network, excluding all the proposed modules. This network has the same backbone as our baseline (VGG16). However, the PVF module is completely removed from the network. Besides, the DPR module is replaced with a sequence of two convolutional layers, each of which is followed by a batch normalization layer and a ReLU layer. The connections between the encoder and decoder remain the same as DeepPyram. Besides, DeepPyram++ in ablation studies is the nested version of DeepPyram implemented based on UNet++.

\begin{figure*}[ht]
\begin{tabular}{cc}
\multicolumn{2}{c}{
\begin{tikzpicture}
    \begin{customlegend}[legend entries={Average,Max,Min}, legend columns=-1]
    \addlegendimage{black,draw=BurntOrange, fill=BurntOrange,mark=*,only marks,very thick,mark options={scale=1.5}}
    \addlegendimage{black,draw=LimeGreen, fill=LimeGreen,mark=square*,only marks,very thick,mark options={scale=1}}
    \addlegendimage{black,draw=BrickRed,fill=BrickRed,mark=square*,only marks,very thick,mark options={scale=1}} 
    \end{customlegend}
\end{tikzpicture}
}\\
\begin{subfigure}{.48\textwidth}\vspace{-0.5\baselineskip}
  \centering
\pgfplotstableread{
x y y-min y-max
{U-Net} 67.91 0 94.76
{SegNet} 83.67 19.15 92.47
{PSPNet+}  88.89 67.67 96.89
{FEDNet}  87.38 61.99 95.85
{RAUNet} 85.34 0 94.79
{CE-Net} 80.43 29.83 93.04
{MultiResUNet} 71.62 4.94 94.93
{dU-Net} 69.46 0 95.83
{PAANet} 84.83 15.64 94.77
{BARNet} 78.85 0 92.82
{CPFNet} 83.74 8.91 95.20
{UNet++\slash DS} 87.85 53.36 96.25
{UNet++}  89.34 64.85 96.87
{DeepPyram}  91.98 55.56 98.07

}{\differanser}
\begin{adjustbox}{height=0.2\textheight}
\begin{tikzpicture}[scale=1.3] 
\begin{axis} [
width  = 0.85\textwidth,
height = 8cm,
symbolic x coords={{U-Net},{SegNet},{PSPNet+},{FEDNet},{RAUNet},{CE-Net},{MultiResUNet},{dU-Net}, {PAANet},{BARNet},{CPFNet},{UNet++\slash DS},{UNet++},{DeepPyram}},
minor ytick={5,10,15,20,25},
yminorgrids,
xtick=data,
ticklabel style = {font=\tiny},
x tick label style={rotate=45,anchor=east},
legend style={at={(0.5,1.25)},anchor=north,legend columns=-1},
ymin=70,ymax=100,
height=.5\textwidth,
ylabel= Dice (\%),
label style={font=\tiny},
tick label style={font=\tiny},
extra y ticks=91.98,
extra y tick labels={},
extra y tick style={
ymajorgrids=true,
ytick style={/pgfplots/major tick length=0pt,
},
grid style={BurntOrange,dashed,/pgfplots/on layer=axis foreground,},
},
]

\addplot+[ForestGreen, very thick, forget plot,only marks,forget plot] 
plot[very thick, error bars/.cd, y dir=plus, y explicit]
table[x=x,y=y,y error expr=\thisrow{y-max}-\thisrow{y}] {\differanser};

\addplot+[red, very thick, only marks,xticklabels=\empty,forget plot] 
plot[very thick, error bars/.cd, y dir=minus, y explicit]
table[x=x,y=y,y error expr=\thisrow{y}-\thisrow{y-min}] {\differanser};

\addplot[only marks,mark=*,mark options={fill=BurntOrange,draw=BurntOrange,very thick}] 
table[x=x,y expr=\thisrow{y}] {\differanser};

\addplot[only marks,mark=square*,color=LimeGreen, mark options={scale=0.8}] 
table[x=x,y expr=\thisrow{y-max}] {\differanser};

\addplot[only marks,mark=square*,color=BrickRed, mark options={scale=0.8}] 
table[x=x,y expr=\thisrow{y-min}] {\differanser};

\end{axis} 
\end{tikzpicture}
\end{adjustbox}
\vspace{-1\baselineskip}
\caption{Lens}
\label{fig:sub-first}
\end{subfigure} &

\begin{subfigure}{.48\textwidth}\vspace{-0.5\baselineskip}
  \centering
\pgfplotstableread{
x y y-min y-max
{U-Net} 89.36 37.62 99.22
{SegNet} 91.31 66.04 95.60
{PSPNet+}  96.45 86.98 98.74
{FEDNet}  96.85 92.42 98.90
{RAUNet} 94.26 83.91 97.34
{CE-Net} 90.89 81.41 97.51
{MultiResUNet} 66.80 2.64 98.68
{dU-Net} 79.53 10.38 98.15
{PAANet} 94.59 73.79 98.31
{BARNet} 95.32 73.75 98.60
{CPFNet} 94.83 87.88 98.54
{UNet++\slash DS} 97.53 87.07 99.32
{UNet++}  97.77 95.08 99.22
{DeepPyram}  98.24 95.28 99.20
}{\differanser}
\begin{adjustbox}{height=0.2\textheight}
\begin{tikzpicture}[scale=1.3] 
\begin{axis} [
width  = 0.85\textwidth,
height = 8cm,
symbolic x coords={{U-Net},{SegNet},{PSPNet+},{FEDNet},{RAUNet},{CE-Net},{MultiResUNet},{dU-Net}, {PAANet},{BARNet},{CPFNet},{UNet++\slash DS},{UNet++},{DeepPyram}},
minor ytick={5,10,15,20,25},
yminorgrids,
xtick=data,
ticklabel style = {font=\tiny},
x tick label style={rotate=45,anchor=east},
legend style={at={(0.5,1.25)},anchor=north,legend columns=-1},
ymin=70,ymax=100,
height=.5\textwidth,
ylabel= Dice (\%),
label style={font=\tiny},
tick label style={font=\tiny},
extra y ticks=98.24,
extra y tick labels={},
extra y tick style={
ymajorgrids=true,
ytick style={/pgfplots/major tick length=0pt,
},
grid style={BurntOrange,dashed,/pgfplots/on layer=axis foreground,},
},
]

\addplot+[ForestGreen, very thick, forget plot,only marks,forget plot] 
plot[very thick, error bars/.cd, y dir=plus, y explicit]
table[x=x,y=y,y error expr=\thisrow{y-max}-\thisrow{y}] {\differanser};

\addplot+[red, very thick, only marks,xticklabels=\empty,forget plot] 
plot[very thick, error bars/.cd, y dir=minus, y explicit]
table[x=x,y=y,y error expr=\thisrow{y}-\thisrow{y-min}] {\differanser};

\addplot[only marks,mark=*,mark options={fill=BurntOrange,draw=BurntOrange,very thick}] 
table[x=x,y expr=\thisrow{y}] {\differanser};

\addplot[only marks,mark=square*,color=LimeGreen, mark options={scale=0.8}] 
table[x=x,y expr=\thisrow{y-max}] {\differanser};

\addplot[only marks,mark=square*,color=BrickRed, mark options={scale=0.8}] 
table[x=x,y expr=\thisrow{y-min}] {\differanser};

\end{axis} 
\end{tikzpicture}
\end{adjustbox}
\vspace{-1\baselineskip}
\caption{Pupil}
\label{fig:sub-first}
\end{subfigure}\\
\begin{subfigure}{0.48\textwidth}\vspace{-0.5\baselineskip}
  \centering
\pgfplotstableread{

x y y-min y-max
{U-Net} 86.20 59.59 96.78
{SegNet} 88.60 70.14 96.10
{PSPNet+}  93.55 81.90 97.86
{FEDNet}  92.10 63.48 97.30
{RAUNet} 92.18 53.21 97.98
{CE-Net} 90.85 77.39 96.97
{MultiResUNet} 83.40 39.51 96.38
{dU-Net} 90.00 65.38 97.36
{PAANet} 92.71 71.63 98.01
{BARNet} 91.09 73.17 97.93
{CPFNet} 93.28 76.78 97.67
{UNet++\slash DS} 92.57 45.19 97.50
{UNet++}  91.72 70.31 97.89
{DeepPyram}  94.63 81.37 98.40

}{\differanser}
\begin{adjustbox}{height=0.2\textheight}
\begin{tikzpicture}[scale=1.3] 
\begin{axis} [
width  = 0.85\textwidth,
height = 8cm,
symbolic x coords={{U-Net},{SegNet},{PSPNet+},{FEDNet},{RAUNet},{CE-Net},{MultiResUNet},{dU-Net}, {PAANet},{BARNet},{CPFNet},{UNet++\slash DS},{UNet++},{DeepPyram}},
minor ytick={5,10,15,20,25},
yminorgrids,
xtick=data,
ticklabel style = {font=\tiny},
x tick label style={rotate=45,anchor=east},
legend style={at={(0.5,1.25)},anchor=north,legend columns=-1},
ymin=70,ymax=100,
height=.5\textwidth,
ylabel= Dice (\%),
label style={font=\tiny},
tick label style={font=\tiny},
extra y ticks=94.63,
extra y tick labels={},
extra y tick style={
ymajorgrids=true,
ytick style={/pgfplots/major tick length=0pt,
},
grid style={BurntOrange,dashed,/pgfplots/on layer=axis foreground,},
},
]

\addplot+[ForestGreen, very thick, forget plot,only marks,forget plot] 
plot[very thick, error bars/.cd, y dir=plus, y explicit]
table[x=x,y=y,y error expr=\thisrow{y-max}-\thisrow{y}] {\differanser};

\addplot+[red, very thick, only marks,xticklabels=\empty,forget plot] 
plot[very thick, error bars/.cd, y dir=minus, y explicit]
table[x=x,y=y,y error expr=\thisrow{y}-\thisrow{y-min}] {\differanser};

\addplot[only marks,mark=*,mark options={fill=BurntOrange,draw=BurntOrange,very thick}] 
table[x=x,y expr=\thisrow{y}] {\differanser};

\addplot[only marks,mark=square*,color=LimeGreen, mark options={scale=0.8}] 
table[x=x,y expr=\thisrow{y-max}] {\differanser};

\addplot[only marks,mark=square*,color=BrickRed, mark options={scale=0.8}] 
table[x=x,y expr=\thisrow{y-min}] {\differanser};

\end{axis} 
\end{tikzpicture}
\end{adjustbox}
\vspace{-1\baselineskip}
\caption{Cornea}
\label{fig:sub-first}
\end{subfigure} &
\begin{subfigure}{.48\textwidth}\vspace{-0.5\baselineskip}
  \centering
\pgfplotstableread{
x y y-min y-max
{U-Net} 67.02 0 98.13
{SegNet} 0 0 0
{PSPNet+}  76.47 0 96.30
{FEDNet}  76.11 0 96.87
{RAUNet} 77.99 0 97.13
{CE-Net} 74.64 1.94 94.15
{MultiResUNet} 66.07 0 97.62
{dU-Net} 71.55 0 97.61
{PAANet} 75.24 0 97.88
{BARNet} 77.14 0 97.50
{CPFNet} 73.51 0 94.95
{UNet++\slash DS} 75.91 0 97.78
{UNet++}  79.56 0 97.81
{DeepPyram}  83.30 0 98.19

}{\differanser}
\begin{adjustbox}{height=0.2\textheight}
\begin{tikzpicture}[scale=1.3] 
\begin{axis} [
width  = 0.85\textwidth,
height = 8cm,
symbolic x coords={{U-Net},{SegNet},{PSPNet+},{FEDNet},{RAUNet},{CE-Net},{MultiResUNet},{dU-Net}, {PAANet},{BARNet},{CPFNet},{UNet++\slash DS},{UNet++},{DeepPyram}},
minor ytick={5,10,15,20,25},
yminorgrids,
xtick=data,
ticklabel style = {font=\tiny},
x tick label style={rotate=45,anchor=east},
legend style={at={(0.5,1.25)},anchor=north,legend columns=-1},
ymin=70,ymax=100,
height=.5\textwidth,
ylabel= Dice (\%),
label style={font=\tiny},
tick label style={font=\tiny},
extra y ticks=83.30,
extra y tick labels={},
extra y tick style={
ymajorgrids=true,
ytick style={/pgfplots/major tick length=0pt,
},
grid style={BurntOrange,dashed,/pgfplots/on layer=axis foreground,},
},
]

\addplot+[ForestGreen, very thick, forget plot,only marks,forget plot] 
plot[very thick, error bars/.cd, y dir=plus, y explicit]
table[x=x,y=y,y error expr=\thisrow{y-max}-\thisrow{y}] {\differanser};

\addplot+[red, very thick, only marks,xticklabels=\empty,forget plot] 
plot[very thick, error bars/.cd, y dir=minus, y explicit]
table[x=x,y=y,y error expr=\thisrow{y}-\thisrow{y-min}] {\differanser};

\addplot[only marks,mark=*,mark options={fill=BurntOrange,draw=BurntOrange,very thick}] 
table[x=x,y expr=\thisrow{y}] {\differanser};

\addplot[only marks,mark=square*,color=LimeGreen, mark options={scale=0.8}] 
table[x=x,y expr=\thisrow{y-max}] {\differanser};

\addplot[only marks,mark=square*,color=BrickRed, mark options={scale=0.8}] 
table[x=x,y expr=\thisrow{y-min}] {\differanser};

\end{axis} 

\end{tikzpicture}
\end{adjustbox}
\vspace{-1\baselineskip}
\caption{Instruments}
\label{fig:sub-first}
\end{subfigure}

\end{tabular}
\caption{Quantitative comparison of segmentation results for the proposed (DeepPyram) and rival architectures (some minimum and average values are not visible due to y-axis clipping).}
\label{fig:dice}
\end{figure*}
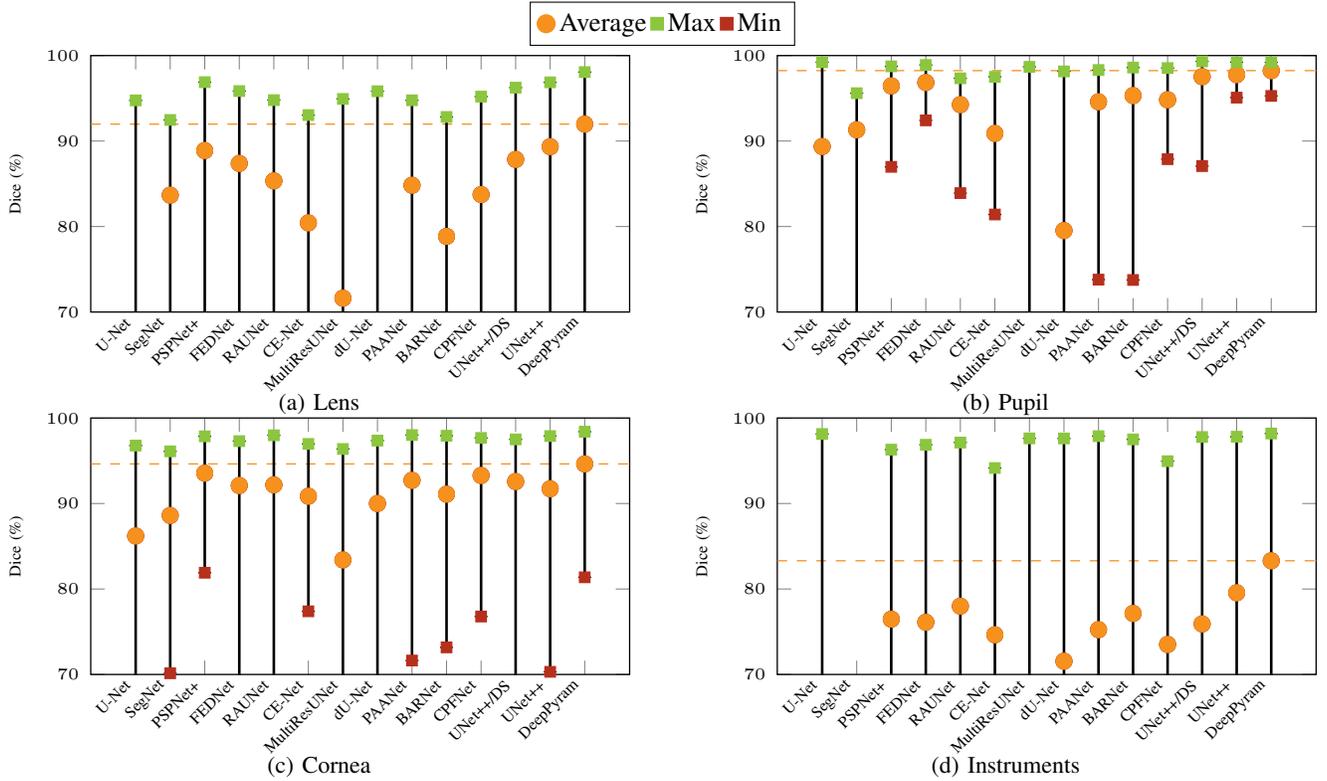
\section{Experimental Results}
\label{sec: Experimental Results}
\subsection{Relevant Object Segmentation}
Fig.~\ref{IoU} compares the resulting IoU of DeepPyram and thirteen rival approaches\footnote{SegNet did not converge during training for instrument segmentation with different initial learning rates.}. Overall, the rival approaches have shown a different level of performance for each of the four relevant objects.
Based on the mean IoU, the best four segmentation approaches for each of the four relevant objects are listed in descending order below:
\begin{itemize}
    \item \textit{Lens:} DeepPyram, Unet++, PSPNet+, UNet++/DS
    \item \textit{Pupil:} DeepPyram, Unet++, UNet++/DS, FEDNet
    \item \textit{Cornea:} DeepPyram, PSPNet+, CPFNet, UNet++/DS
    \item \textit{Instrument:} DeepPyram, Unet++, RAUNet, BARNet
\end{itemize}
Accordingly, DeepPyram, Unet++, and PSPNet+ contribute to the top three segmentation results for the relevant objects in cataract surgery videos. However, DeepPyram shows considerable improvement in segmentation accuracy compared to the second-best approach in each class. Specifically, DeepPyram has achieved more than $4\%$ improvement in lens segmentation ($85.61\%$ vs. $81.32\%$) and more than $4\%$ improvement in instrument segmentation ($74.40\%$ vs. $70.11\%$) compared to UNet++ as the second-best approach. Moreover, DeepPyram appears to be the most reliable approach considering the smallest standard deviation compared to the rival approaches. This significant improvement is attributed to the PVF, DPR, and $P\mathcal{L}$ modules. 

As shown in Fig~\ref{fig:dice}, DeepPyram has achieved the highest dice coefficient compared to the rival approaches for the lens, pupil, cornea, and instrument segmentation. Moreover, DeepPyram is the most reliable segmentation approach based on achieving the highest minimum dice percentage.

\begin{figure}[!tb]
    \centering
    \includegraphics[width=0.94\columnwidth]{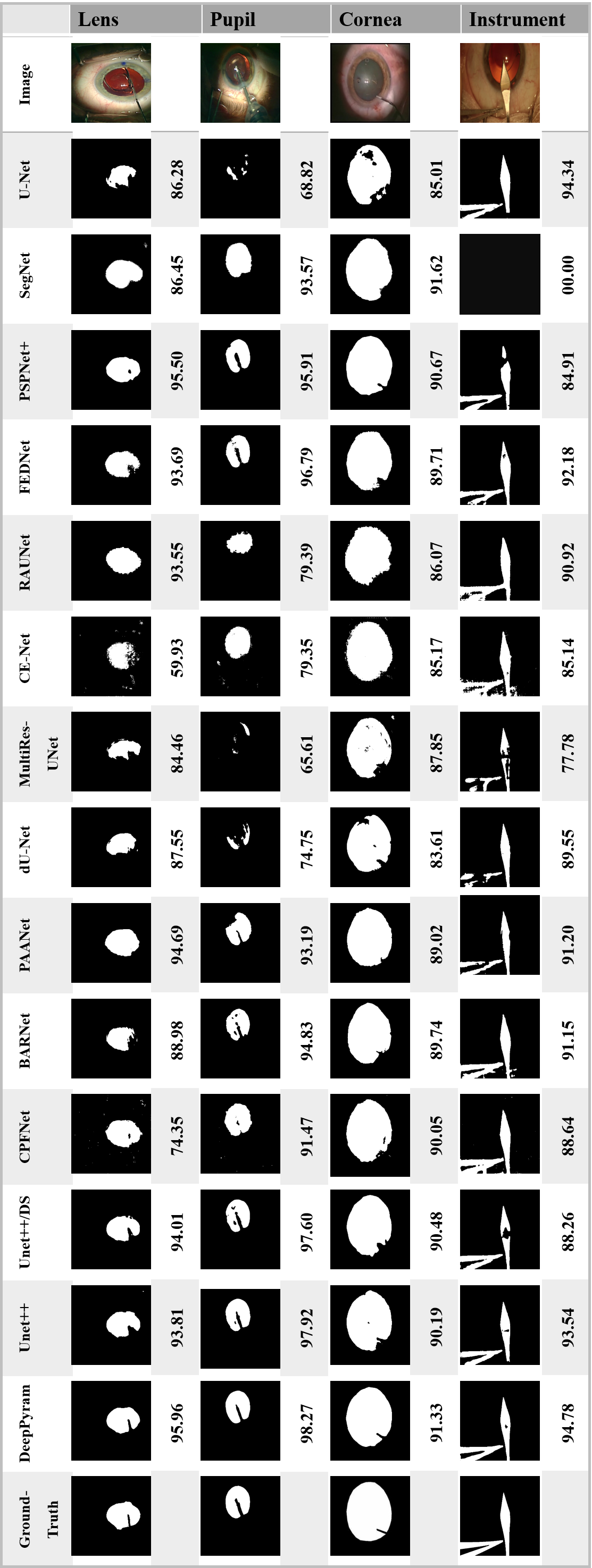}
    \caption{Qualitative comparisons among DeepPyram and the rival approaches for the relevant objects in cataract surgery videos (the numbers denote the Dice(\%) coefficient for each detection). The representative images are selected from the test set.}
    \label{fig:Vis}
\end{figure}

Fig.~\ref{fig:Vis} further affirms the effectiveness of DeepPyram in enhancing the segmentation results. Taking advantage of the pyramid view provided by the PVF module, DeepPyram can handle reflection and brightness variation in instruments, blunt edges in the cornea, color and texture variation in the pupil, as well as transparency in the lens. Furthermore, powering by deformable pyramid reception, DeepPyram can tackle scale variations in instruments and blunt edges in the cornea. In particular, we can perceive from Fig.~\ref{fig:Vis} that DeepPyram shows much less distortion in the region of edges (especially in the case of the cornea. Furthermore, based on these qualitative experiments, DeepPyram shows much better precision and recall in the narrow regions for segmenting the instruments and other relevant objects in the case of occlusion by the instruments.

\subsection{Ablation Study}

Table~\ref{tab:modules} validates the effectiveness of the proposed modules in segmentation enhancement. The PVF module can notably enhance the performance for cornea and instrument segmentation ($2.41\%$ and $2.76\%$ improvement in IoU, respectively). This improvement is due to the ability of the PVF module to provide a global view of varying-size sub-regions centering around each distinctive spatial position. Such a global view can reinforce semantic representation in the regions corresponding to blunt edges and reflections. Due to scale variance in instruments, the DPR module can effectively boost the segmentation performance for instruments. The addition of $P\mathcal{L}$ module results in the improvement of IoU for all relevant segments, especially lens segmentation (around $2\%$ improvement) and Instrument ($1.64\%$ improvement). The combination of PVF, DPR, and $P\mathcal{L}$ modules can contribute to $4.58\%$ improvement in instrument segmentation and $4.22\%$ improvement in cornea segmentation (based on IoU\%). These modules have improved the IoU for the lens and pupil by $2.85\%$ and $1.43\%$, respectively.

Overall\footnote{The ``Overall" column in Table~\ref{tab:modules} is the mean of the other four average values.}, the addition of different modules of DeepPyram has led to considerable improvement of segmentation performance ($3.27\%$ improvement in IoU) on average compared to the baseline approach. The PVF module can provide varying-angle global information centering around each pixel in the convolutional feature map to support relative information access. The DPR module enables large-field content-adaptive reception while minimizing the additive trainable parameters. 

It should be noted that even our baseline approach has much better performance compared to some rival approaches. We argue that the fusion modules adopted in some rival approaches may lead to the dilution of discriminative semantic information.

\begin{figure*}[tb!]
    \centering
    \includegraphics[width=0.9\textwidth]{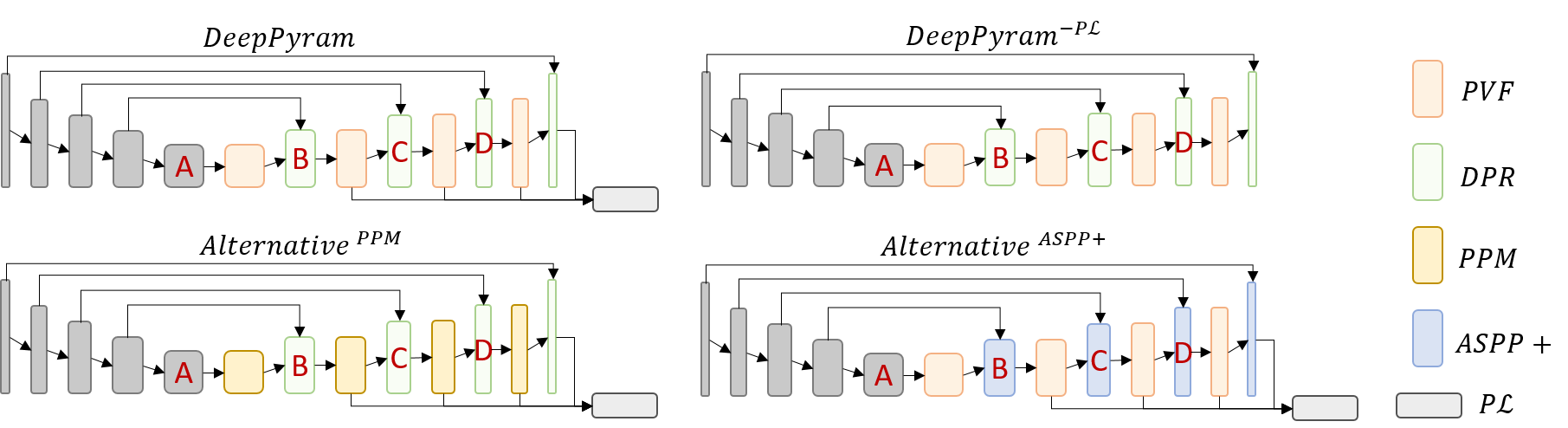}
    \caption{The overall architecture of DeepPyram compared to its three alternatives. The locations A, B, C, and D in each architecture correspond to the four modules for which we visualize the feature representations in Fig.~\ref{fig: replace_module_vis}.}
    \label{fig: replace_module}
\end{figure*}

\begin{figure*}[tb!]
    \centering
    \includegraphics[width=0.9\textwidth]{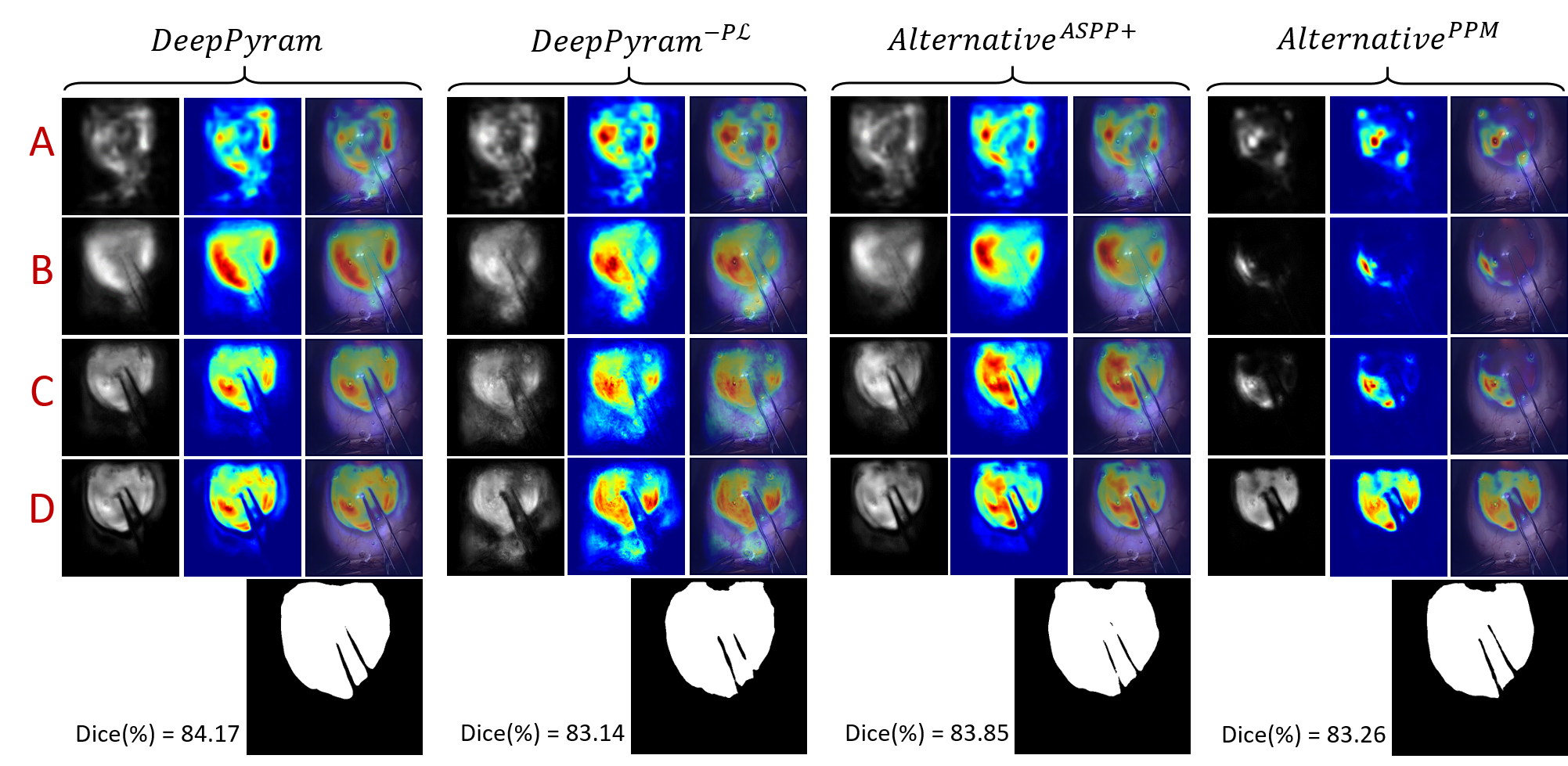}
    \caption{Visualization of the effect of the proposed and alternative modules based on class activation maps~\cite{Score-CAM} using the network architectures demonstrated in Fig.~\ref{fig: replace_module}. For each approach, the figures from left to right represent the gray-scale activation maps, heatmaps, and heatmaps on images for a representative input image from the test set.}
    \label{fig: replace_module_vis}
\end{figure*} 

\subsection{Comparisons with Alternative Modules} 
Herein, we compare the effectiveness of the proposed modules with our enhanced version of the ASPP module~\cite{DeepLab} and PPM~\cite{PSPNet}. More concretely, we replace the PVF module with PPM (referring to as $Alternative^{PPM}$) and DPR module with our improved version of the ASPP module (referring to as $Alternative^{ASPP+}$). Fig.~\ref{fig: replace_module} demonstrates the architecture of DeepPyram versus the alternative networks. The difference between the modules ASPP+ and ASPP lies in the filter size of the convolutional layers. In the ASPP module, there are four parallel convolutional layers: a pixel-wise convolutional layer and three dilated convolutions with different dilation rates. In ASPP+, the pixel-wise convolution is replaced with a $3\times 3$ convolutional layer that effectively enhances the segmentation performance. Instead of three dilated convolutions in ASPP+, we use two dilated convolutions with the same dilation rates as in the DPR module. Moreover, the parallel feature-maps in the ASPP module are fused using a pixel-wise convolution, whereas the ASPP+ module adopts a kernel-size of $3\times 3$ to boost the segmentation performance.   Besides, we have removed the $P\mathcal{L}$ module and referred to it as $DeepPyram^{-P\mathcal{L}}$, to qualitatively compare its performance with DeepPyram. As illustrated in Fig.~\ref{fig: replace_module}, location ``A" corresponds to the output of the last encoder's layer in the bottleneck. locations ``B-D" are the outputs of three modules in the same locations of the decoder networks in DeepPyram and the three alternative networks. Fig.~\ref{fig: replace_module_vis} compares the class activation maps corresponding to these four locations in DeepPyram and alternative approaches for cornea segmentation in a representative image~\footnote{These activation maps are obtained using Score-CAM~\cite{Score-CAM} visualization approach.}. A comparison between the activation maps of DeepPyram and $DeepPyram^{-P\mathcal{L}}$ indicates how negatively removing the $P\mathcal{L}$ module affects the discrimination ability in different semantic levels. It is evident that the activation map of block ``C" in DeepPyram is even more concrete compared to the activation map of block ``D" (which is in a higher semantic level) in $DeepPyram^{-P\mathcal{L}}$. We can infer that the $P\mathcal{L}$ module can effectively reinforce the semantic representations in different semantic levels of the network. The effect of pixel-wise global view (PVF module) versus region-wise global view (PPM) can be inferred by comparing the activation maps of DeepPyram and $Alternative^{PPM}$. The activation maps of $Alternative^{PPM}$ are impaired and distorted in different regions, especially in the lower semantic layers. The activation maps of $Alternative^{ASPP+}$ compared to DeepPyram confirm that replacing the deformable convolutions with regular convolutions can negatively affect semantic representation in narrow regions and the object borders. Table~\ref{tab:module-comparison} provides quantitative comparisons among the proposed and alternative networks. These results further confirm the effectiveness of the proposed PVF and DPR modules compared to their alternatives.

\begin{table*}[!t]
\renewcommand{\arraystretch}{0.9}
\caption{Impact of alternative modules on the segmentation results (IoU\% and Dice\%) of DeepPyram.}
\label{tab:module-comparison}
\centering
\begin{tabular}{lccccccccccc}
\specialrule{.12em}{.05em}{.05em}%
&&\multicolumn{2}{c}{Lens}&\multicolumn{2}{c}{Pupil}&\multicolumn{2}{c}{Cornea}&\multicolumn{2}{c}{Instrument}&\multicolumn{2}{c}{Overall}\\\cmidrule(lr){3-4}\cmidrule(lr){5-6}\cmidrule(lr){7-8}\cmidrule(lr){9-10}\cmidrule(lr){11-12}
Network&Params&IoU(\%)&Dice(\%)&IoU(\%)&Dice(\%)&IoU(\%)&Dice(\%)&IoU(\%)&Dice(\%)&IoU(\%)&Dice(\%)\\\specialrule{.12em}{.05em}{.05em}%
$Alternative^{ASPP+}$&22.99 M&85.02&91.51&96.41&98.17&88.83&94.00&\textbf{74.81}&\textbf{83.66}&86.26&91.83\\
$Alternative^{PPM}$&23.44 M&83.40&90.66&95.70&97.79&87.44&92.88&72.51&81.03&84.76&90.59\\
DeepPyram&23.62 M&\textbf{85.84}&\textbf{91.98}&\textbf{96.56}&\textbf{98.24}&\textbf{90.24}&\textbf{94.77}&74.40&83.30&\textbf{86.76}&\textbf{92.07}\\\specialrule{.12em}{.05em}{.05em}%

\end{tabular}
\label{tab:alternative}
\end{table*}

\begin{table*}[!t]
\renewcommand{\arraystretch}{0.9}
\caption{Impact of different backbones and combinations on the segmentation results (IoU\% and Dice\%) of DeepPyram.}
\label{tab:ablation}
\centering
\begin{tabular}{llccccccccccc}
\specialrule{.12em}{.05em}{.05em}%
&&&\multicolumn{2}{c}{Lens}&\multicolumn{2}{c}{Pupil}&\multicolumn{2}{c}{Cornea}&\multicolumn{2}{c}{Instrument}&\multicolumn{2}{c}{Overall}\\\cmidrule(lr){4-5}\cmidrule(lr){6-7}\cmidrule(lr){8-9}\cmidrule(lr){10-11}\cmidrule(lr){12-13}
&&Params&IoU(\%)&Dice(\%)&IoU(\%)&Dice(\%)&IoU(\%)&Dice(\%)&IoU(\%)&Dice(\%)&IoU(\%)&Dice(\%)\\\specialrule{.12em}{.05em}{.05em}%
\multicolumn{1}{ c|  }{\parbox[t]{1mm}{\multirow{4}{*}{\rotatebox[origin=c]{90}{Backbone}}}}&ResNet50&85.52 M&81.71&89.54&95.09&97.46&89.32&94.23&71.79&80.73&84.48&90.49\\
\multicolumn{1}{ c|  }{}&ResNet34&25.77 M&82.77&90.22&95.06&97.45&88.68&93.84&72.58&80.99&84.77&90.62\\
\multicolumn{1}{ c|  }{}&VGG19&28.93 M&85.33&91.66&96.36&98.14&88.77&93.93&\textbf{74.70}&\textbf{83.49}&86.29&91.80\\
\multicolumn{1}{ c|  }{}&VGG16&23.62 M&\textbf{85.84}&\textbf{91.98}&\textbf{96.56}&\textbf{98.24}&\textbf{90.24}&\textbf{94.77}&74.40&83.30&\textbf{86.76}&\textbf{92.07}\\
\hline\hline
\multicolumn{2}{ c }{DeepPyram++}&28.48 M&84.83&91.54&96.30&98.11&89.48&94.34&74.64&83.34&86.31&91.83\\\specialrule{.12em}{.05em}{.05em}%

\end{tabular}
\label{tab:backbone}
\end{table*}

\begin{table*}[!t]
\renewcommand{\arraystretch}{0.9}
\caption{Impact of different super-resolution functions on the segmentation results (IoU\% and Dice\%) of DeepPyram.}
\label{tab:ablation}
\centering
\begin{tabular}{lccccccccccc}
\specialrule{.12em}{.05em}{.05em}%
&&\multicolumn{2}{c}{Lens}&\multicolumn{2}{c}{Pupil}&\multicolumn{2}{c}{Cornea}&\multicolumn{2}{c}{Instrument}&\multicolumn{2}{c}{Overall}\\\cmidrule(lr){3-4}\cmidrule(lr){5-6}\cmidrule(lr){7-8}\cmidrule(lr){9-10}\cmidrule(lr){11-12}
Upsampling&Params&IoU(\%)&Dice(\%)&IoU(\%)&Dice(\%)&IoU(\%)&Dice(\%)&IoU(\%)&Dice(\%)&IoU(\%)&Dice(\%)\\\specialrule{.12em}{.05em}{.05em}%
Trans Conv&25.01 M&84.37&91.22&96.01&97.96&88.80&93.97&\textbf{75.16}&\textbf{83.71}&86.08&91.71\\
PixelShuffle&36.15 M&84.31&91.00&96.49&98.20&89.17&94.15&74.55&83.41&86.13&91.69\\
Bilinear&23.62 M&\textbf{85.84}&\textbf{91.98}&\textbf{96.56}&\textbf{98.24}&\textbf{90.24}&\textbf{94.77}&74.40&83.30&\textbf{86.76}&\textbf{92.07}\\\specialrule{.12em}{.05em}{.05em}%

\end{tabular}
\label{tab:upsampling}
\end{table*}

\subsection{Effect of Different Backbones and Nested Architecture}

We have evaluated the achievable segmentation accuracy using different backbone networks to decide which backbone performs the best, considering the trade-off between the number of trainable parameters and the Dice percentage. As listed in Table~\ref{tab:backbone}, the experimental results show that the higher-depth networks such as ResNet50 cannot improve the segmentation accuracy. In contrast, VGG16 with the fewest number of trainable parameters has achieved the best segmentation performance. Moreover, VGG19, having around 5.3M parameters more than VGG16, performs just slightly better than the baseline backbone in instrument segmentation. In Table~\ref{tab:backbone}, DeepPyram++ is the nested version of DeepPyram \footnote{UNet++ uses the encoder-decoder architecture of U-Net as baseline, and adds additional convolutional layers between different encoder's and decoder's layers to form a nested multi-depth architecture. In DeepPyram++, we replace the the encoder-decoder baseline of UNet++ (which is U-Net) with our proposed DeepPyram network to see if these additional layers and connections can improve the segmentation performance of DeepPyram.} (with the same connections as in UNet++). This nested architecture shows around a one percent drop in IoU percentage on average. 

\subsection{Effect of Different Super-resolution Functions}

Table~\ref{tab:upsampling} compares the effect of three different super-resolution functions on the segmentation performance, including transposed convolution, Pixel-Shuffle~\cite{PixelShuffle}, and bilinear upsampling. Overall, the network with bilinear upsampling function with the fewest parameters has achieved the best performance among the networks with different upsampling functions. Besides, the results reveal that the bilinear upsampling function has the best performance in segmenting all relevant objects except for the instruments. 

\section{Conclusion}
\label{sec: Conclusion}
In recent years, considerable attention has been devoted to computerized surgical workflow analysis for various applications such as action recognition, irregularity detection, objective skill assessment, and so forth. A reliable relevant-instance-segmentation approach is a prerequisite for a majority of these applications. In this paper, we have proposed a novel network architecture for semantic segmentation in cataract surgery videos. The proposed architecture takes advantage of three modules, namely ``Pyramid View Fusion", ``Deformable Pyramid Reception", and ``Pyramid Loss",  to simultaneously deal with different challenges. These challenges include: (i) geometric transformations such as scale variation and deformability, (ii) blur degradation and blunt edges, and (iii) transparency, and texture and color variation. Experimental results have shown the effectiveness of the proposed network architecture (DeepPyram) in retrieving the object information in all mentioned situations. DeepPyram stands in the first position for cornea, pupil, lens, and instrument segmentation compared to all rival approaches. The proposed architecture can also be adopted for various other medical image segmentation and general semantic segmentation problems.

\balance
\bibliographystyle{IEEEtran}
\bibliography{bibtex.bib}

\end{document}